\documentclass[11pt]{article}
\usepackage[a4paper,margin=1in]{geometry}

\usepackage{amsmath,amsfonts,bm}

\def\eqref#1{equation~\ref{#1}}

\def\1{\bm{1}}

\DeclareMathAlphabet{\mathsfit}{\encodingdefault}{\sfdefault}{m}{sl}
\SetMathAlphabet{\mathsfit}{bold}{\encodingdefault}{\sfdefault}{bx}{n}

\usepackage{hyperref}
\usepackage{url}
\usepackage{graphicx}
\usepackage{booktabs}
\usepackage{makecell}
\usepackage[most]{tcolorbox}

\newcommand{\database}[0]{GVRS}
\usepackage{wrapfig}
\usepackage{enumitem}
\usepackage{multirow}
\usepackage{authblk}

\title{H-GRPO: Permutation-Invariant Reinforcement Learning for Improving Spatial Grounded Reasoning}

\author[1,2]{Eric Peh}
\author[3]{Debaditya Roy}
\author[1,2]{Chen Li}
\author[1,2,4]{Basura Fernando}

\affil[1]{Institute of Advanced Intelligence and Computing (IAIC), Agency for Science, Technology and Research, Singapore.}
\affil[2]{Centre for Frontier AI Research, Agency for Science, Technology and Research, Singapore.}
\affil[3]{Department of Computer Science and Engineering, Indian Institute of Technology Kharagpur, India.}
\affil[4]{College of Computing and Data Science, Nanyang Technological University, Singapore.}

\date{}

\begin{document}

\maketitle

\begin{abstract}
Vision-language models (VLMs) have achieved strong performance on visual reasoning tasks, yet a correct final answer does not necessarily imply that the model followed a correct or visually grounded reasoning process. We study how VLMs can be trained to produce explicit intermediate reasoning steps that are both semantically meaningful and grounded in visual evidence. We represent each reasoning step as a structured triplet consisting of a \emph{sub-question}, \emph{sub-answer}, and \emph{supporting bounding box}, enabling the reasoning process to be evaluated along both semantic and spatial dimensions. To optimize this representation, we introduce \textbf{Hungarian-GRPO (H-GRPO)}, a reinforcement-learning framework that aligns predicted and reference reasoning steps using permutation-invariant Hungarian matching, avoiding the assumption that valid reasoning steps must follow a fixed ordering. 
Experiments show that optimizing grounded reasoning improves both intermediate reasoning quality and answer accuracy, with a H-GRPO variant achieving the strongest overall grounding and decomposition performance. Our analyses show that grounding quality and answer correctness are related but distinct, while counterfactual evidence masking confirms that the localized regions contain information important to the model's predictions. These findings motivate evaluating VLMs beyond answer accuracy and demonstrate the value of explicit, spatially grounded reasoning.
\end{abstract}
\section{Introduction}

\begin{figure}[ht]
    \centering
    \includegraphics[width=\textwidth]{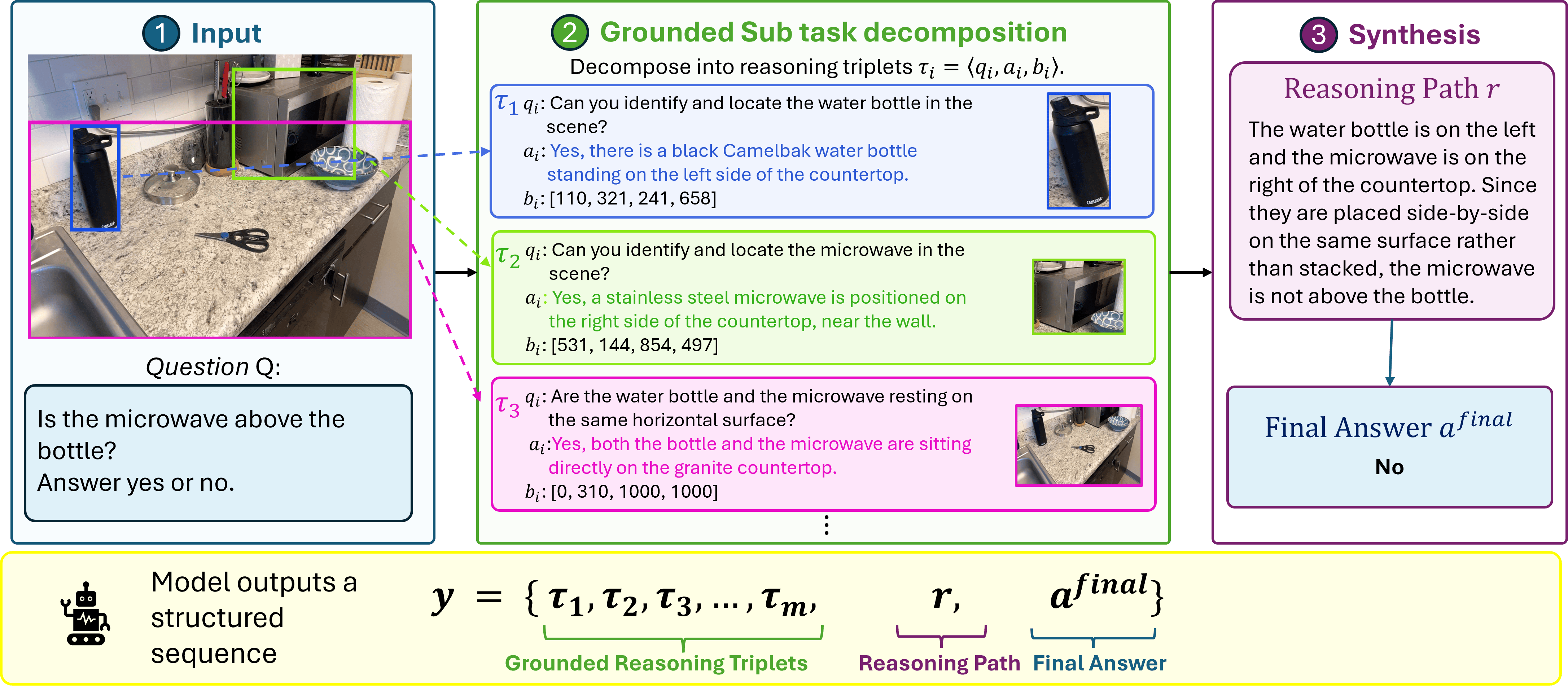}    
    \caption{Overview of our grounded reasoning. Given an image I and question $Q$, the VLM decomposes the reasoning into a sequence of  triplets $\tau_i = \langle q_i, a_i, b_i \rangle$, where each triplet contains a sub-question, intermediate answer, and supporting spatial evidence. The intermediate triplets are subsequently synthesized into a coherent reasoning path $r$ and final answer $a^{final}$, yielding a structured and verifiable reasoning process.}    
    \label{fig:Intro}
\end{figure}
Visual geometric and spatial reasoning is a fundamental aspect of multimodal reasoning. Vision-language models (VLMs) have achieved strong performance on visual question-answering benchmarks that require geometric and spatial reasoning. Previous works have suggested that high answer accuracy does not necessarily indicate faithful visual reasoning~\cite{xiao2024can,shrestha2020negative} and  that models may arrive at the correct answer by exploiting spurious correlations~\cite{shrestha2020negative},   
relying on common-sense knowledge, or using other inference strategies~\cite{turpin2023language} without grounding its reasoning in the relevant visual evidence~\cite{reich2023measuring}. This gap is particularly concerning when VLMs are post-trained with reinforcement learning (RL), as optimizing solely for final-answer correctness may improve answer accuracy without improving evidence-grounded reasoning.


Recent RL-based multimodal reasoning methods, such as R1-VL~\cite{zhang2025r1vl}, Vision-R1~\cite{huang2025vision}, Reason-RFT~\cite{tan2025reasonrft}, and VLM-R1~\cite{shen2025vlmr1}, improve VLM reasoning through outcome-level or process-oriented rewards. However, intermediate reasoning is predominantly represented and optimized in the textual space, without explicitly requiring each reasoning step to be grounded in its supporting visual evidence. Consequently, a plausible intermediate reasoning step does not necessarily indicate that the model has attended to or localized the relevant image region. Grounding-oriented methods such as Visual-RFT~\cite{liu2025visualrft} and ViGoRL~\cite{sarch2025vigorl} incorporate localization into RL training, but do not explicitly structure each intermediate reasoning step as a coupled \emph{sub-question, sub-answer, and localized visual evidence} tuple. Similarly, datasets such as Visual CoT~\cite{shao2024visual} provide visual chain-of-thought supervision with visual evidence, but do not generally associate every intermediate reasoning step with its corresponding localized evidence. This leaves a gap in jointly supervising \emph{what} intermediate question should be resolved, \emph{what} intermediate conclusion should be reached, and \emph{where} the supporting evidence lies in the image. As illustrated in Figure~\ref{fig:Intro}, we address this gap by explicitly decomposing visual reasoning into grounded intermediate steps and optimizing both their semantic reasoning and step-specific spatial grounding.

To optimize such grounded reasoning processes, we introduce a reinforcement learning objective that jointly rewards final-answer correctness, semantic quality of the intermediate reasoning, and spatial alignment of the associated visual evidence. A key challenge is that a reasoning problem may admit multiple valid intermediate decompositions whose steps differ in order while expressing equivalent reasoning. We therefore formulate process-level reward using step matching between predicted and reference reasoning trajectories, allowing semantically corresponding steps and their visual evidence to be aligned before computing the reward. This enables the model to learn from grounded reasoning trajectories without requiring a fixed correspondence between predicted and reference step order, while retaining final-answer correctness as the outcome-level objective.

We choose QA decomposition as a practical intermediate representation because it simultaneously expresses the reasoning target, its inferred answer, and its supporting visual evidence in a form that is directly evaluable by existing VLMs and amenable to reinforcement learning. In contrast to free-form chain-of-thought~\cite{shao2024visual}, each step has a clear role: the sub-question specifies what needs to be established, the sub-answer records the inferred fact, and the bounding box identifies the visual evidence supporting that inference. This structure enables intermediate reasoning steps to be evaluated using both semantic and spatial criteria. We extend considered  spatial-relation benchmarks with annotations and metrics for evaluating such grounded step-by-step reasoning.

While this structured representation enables step-level supervision, it also raises a key challenge: \emph{how should a predicted reasoning decomposition be compared with a reference when multiple valid reasoning paths exist?} Different models may reach the same answer through different intermediate steps or express equivalent reasoning steps in different orders. A fixed sequential correspondence between predicted and reference steps may be appropriate for some reasoning paths, but it can also unfairly penalize valid decompositions that express equivalent reasoning in a different order. To address this issue, we introduce \textbf{Hungarian Group Relative Policy Optimization (H-GRPO)}, which uses bipartite matching to align predicted and reference reasoning steps before computing process-level rewards. For settings where ordered supervision may provide a useful initialization, H-GRPO+ first optimizes the model with fixed step matching and subsequently applies permutation-invariant H-GRPO. This two-stage procedure first encourages structured, order-sensitive reasoning trajectories and then relaxes the ordering constraint to accommodate valid alternative step orderings. When the matching is restricted to the identity assignment, such that each predicted step is paired with the reference step at the same position, H-GRPO reduces to the corresponding sequential GRPO-style objective.

Our framework consists of three main components. First, we introduce a \textbf{grounded reasoning decomposition protocol} in which each intermediate step contains a sub-question, sub-answer, and supporting evidence bounding box. Second, we develop a \textbf{data synthesis pipeline} for generating step-wise grounded reasoning traces with limited manual annotation. Third, we introduce \textbf{H-GRPO}, which uses Hungarian matching to provide permutation-invariant process-level supervision and complements standard GRPO when reasoning order should be preserved.

Beyond training, we introduce a unified evaluation protocol that considers final-answer correctness together with spatial grounding, step-wise reasoning alignment, and reliance on visual evidence. Our experiments examine how method variants balance reasoning and grounding, and whether explicitly supervising grounded intermediate reasoning can improve these dimensions jointly.

In summary, our main contribution is H-GRPO, a process-level reinforcement learning framework for grounded visual reasoning. H-GRPO represents intermediate reasoning as sub-question--sub-answer--evidence triplets and introduces permutation-invariant matching to align predicted and reference reasoning steps based on their semantic and spatial correspondence. This enables direct optimization of both reasoning correctness and visual grounding while allowing valid variations in the ordering of intermediate reasoning steps.
\section{Related Work}
Chain-of-thought reasoning has been extended to VLMs for compositional and multi-step visual reasoning. Visual CoT~\cite{shao2024visual} and Multimodal-CoT~\cite{zhang2024multimodal} introduce intermediate reasoning over visual-textual inputs, while ScienceQA~\cite{lu2022learn} and Kosmos-2~\cite{peng2024kosmos} support multimodal science reasoning and visual grounding. Recent work further explores step-wise supervision, self-reflection, and tool-assisted reasoning~\cite{zhang2025r1vl,huang2025vision,tan2025reasonrft,wang2025vlrethinker}. However, most visual CoT methods supervise reasoning mainly in language space, without explicitly grounding each reasoning step in localized image evidence, allowing plausible but visually unsupported chains.

Reward-based post-training has been used to improve VLM factuality and alignment. LLaVA-RLHF~\cite{sun2024aligning} uses Factually Augmented RLHF with caption-based factual signals to reduce hallucination, while RLHF-V~\cite{yu2024rlhf} improves data efficiency through segment-level corrections and Dense DPO. Related RL methods with verifiable rewards, including Visual-RFT~\cite{liu2025visualrft} based on DeepSeek-R1~\cite{guo2025deepseek} and VLM-R1~\cite{shen2025vlmr1}, optimize perception and grounding using rewards such as IoU or accuracy. However, these approaches mainly target preference alignment, factuality, or task-level grounding, rather than enforcing step-wise visual grounding throughout the reasoning process.

Motivated by DeepSeek-R1~\cite{guo2025deepseek}, recent work applies RL to multi-step visual reasoning. Vision-R1~\cite{huang2025vision} combines cold-start multimodal CoT with GRPO, R1-VL~\cite{zhang2025r1vl} introduces StepGRPO for intermediate-step rewards, Reason-RFT~\cite{tan2025reasonrft} uses SFT-then-GRPO for robust cross-domain reasoning, and VL-Rethinker~\cite{wang2025vlrethinker} mitigates vanishing advantages via selective replay and forced rethinking. However, these methods mainly assess textual reasoning traces or final answers, without enforcing grounding of each reasoning step in image regions.

Recent methods increasingly couple visual evidence with intermediate reasoning. Grounded CoT~\cite{wu2026grounded} and its bootstrapped variant~\cite{xia2025bootstrapping} associate reasoning steps with localized evidence, while ViGoRL~\cite{sarch2025vigorl} uses MCTS-generated grounded traces and multi-turn GRPO with dynamic zooming. Ground-R1~\cite{cao2025ground} learns evidence-grounded reasoning through RL, VLM-R$^3$~\cite{jiang2026vlm} introduces region-conditioned policy optimization for actively acquiring visual evidence, and Vision-SR1~\cite{li2025visionsr1} separates perception from reasoning and rewards intermediate perception quality. While these approaches establish the importance of grounding during visual reasoning, H-GRPO instead represents each reasoning step as an explicit sub-question--sub-answer--evidence triplet and directly rewards its semantic and spatial correspondence to a reference decomposition. Moreover, permutation-invariant Hungarian matching accommodates alternative reasoning orders while preserving step-level visual grounding.



\section{Method}

\paragraph{Overview} We propose \textbf{Hungarian Group Relative Policy Optimization (H-GRPO)}, a fine-tuning framework for grounded visual reasoning. Given an image and a question, the model is trained to produce not only a final answer, but also an explicit decomposition of the reasoning process into grounded intermediate steps. 
Each intermediate step contains a sub-question, an intermediate answer, and a spatial evidence bounding box. 
The key idea is to train the model with a reward that jointly encourages: 
(i) correct output format, 
(ii) correct final answer, and 
(iii) visually grounded and semantically meaningful reasoning decomposition. 
Unlike standard sequence-level reward formulations, our method evaluates intermediate reasoning steps using a permutation-invariant Hungarian matching score. 
This allows different but valid reasoning orders to receive appropriate credit.

\paragraph{Grounded Reasoning Decomposition.}
We require the VLM to transform a monolithic visual reasoning process into a sequence of grounded sub-tasks. 
Given an image $I$ and a question $Q$, the model generates a decomposition 
$
D = \{\tau_1, \tau_2, \dots, \tau_m\},
$
where each reasoning triplet is defined as
$
\tau_i = \langle q_i, a_i, b_i \rangle.
$
Here, $q_i$ denotes a sub-question corresponding to a discrete reasoning step, $a_i$ denotes the model's intermediate answer to that sub-question, and $b_i$ denotes the spatial evidence used to justify the answer. The bounding box is represented as
$
b_i = [x_{\min}, y_{\min}, x_{\max}, y_{\max}].
$
After producing the intermediate reasoning triplets, the VLM synthesizes them into two final components:
\textbf{Reasoning path} ($r$) -- a natural-language explanation that integrates the intermediate evidence $\tau_{1:m}$ into a coherent argument, and \textbf{Final answer} ($a^{\mathrm{final}}$) -- a concise answer to the original question.
Thus, the complete model output is
\[
y = \{\tau_1, \dots, \tau_m, r, a^{\mathrm{final}}\}.
\]
This decomposition encourages the model to expose the visual evidence and steps that lead to the final answer. 

\paragraph{Reinforcement Learning Formulation.}
We formulate grounded visual reasoning as a reinforcement learning problem. 
Let $\pi_\theta$ denote the policy model. 
Given an input image-question pair $(I, Q)$, the policy generates an output trajectory $y$ consisting of grounded reasoning triplets, a reasoning path, and a final answer.
For each prompt, the policy $\pi_\theta$ samples a group of $G$ reasoning trajectories through independent rollouts:
\[
\{y_1, y_2, \dots, y_G\} \sim \pi_\theta(\cdot \mid I, Q).
\]
The policy is optimized to maximize a composite reward $\mathcal{J}$, which evaluates both the final correctness and the quality of the intermediate reasoning process.
Recent GRPO and RLVR methods have shown strong results in domains such as mathematics and code generation, where correctness can often be verified with relatively dense and well-structured rewards~\cite{shao2024deepseekmath, guo2025deepseek}. 
However, spatial and geometric reasoning is more flexible: a question may admit multiple valid reasoning paths that differ in order, structure, or visual grounding. 
A fixed sequence-level comparison between generated and reference reasoning chains may therefore penalize valid reasoning simply because it appears in a different order. 
To address this limitation, we introduce H-GRPO, which replaces order-sensitive reasoning-chain comparison with a permutation-invariant Hungarian matching reward. 

\textbf{Hungarian Matching for Reasoning Evaluation.}
For each generated trajectory $y$, let
$
\hat{D} = \{\hat{\tau}_1, \hat{\tau}_2, \dots, \hat{\tau}_m\}
$
be the set of predicted reasoning triplets, where
\(
\hat{\tau}_i = (\hat{q}_i, \hat{a}_i, \hat{b}_i),
\) and
$
D^* = \{\tau_1^*, \tau_2^*, \dots, \tau_n^*\}
$
be the set of reference reasoning triplets, where
\(
\tau_j^* = (q_j^*, a_j^*, b_j^*).
\)
Instead of comparing predicted and reference triplets in a fixed order, we construct a bipartite matching problem between $\hat{D}$ and $D^*$. 
We first define a similarity matrix $S \in \mathbb{R}^{m \times n}$ where each entry $S_{ij}$ measures the similarity between predicted triplet $\hat{\tau}_i$ and reference triplet $\tau_j^*$. 
The similarity score is defined as
\begin{equation}
S_{ij}=\frac{\mathrm{sim}_q(\hat{q}_i, q_j^*) +\mathrm{sim}_a(\hat{a}_i, a_j^*)+\mathrm{IoU}(\hat{b}_i, b_j^*)\,e_j^{\mathrm{ref}}}{2+e_j^{\mathrm{ref}}} 
\end{equation}
Here, \(e_j^{\mathrm{ref}}\in\{0,1\}\) indicates whether the \(j\)-th reference reasoning step contains a bounding-box annotation. When \(e_j^{\mathrm{ref}}=1\), spatial consistency is included through the IoU term. When \(e_j^{\mathrm{ref}}=0\), the spatial term is omitted and the similarity is determined solely from the sub-question and sub-answer similarities that also includes the case when the prediction contains a bounding box but the reference does not. The terms $\mathrm{sim}_q$ and $\mathrm{sim}_a$ are Sentence-BERT~\cite{reimers2019sentence} cosine similarities between predicted and reference sub-questions and answers. It should be noted that similarity scores are practically always positive and less than one in our context. Therefore, we did not use extra normalization for similarities.
The term $\mathrm{IoU}(\hat{b}_i, b_j^*)$ measures the spatial overlap between predicted and reference bounding boxes.

To handle different numbers of predicted and reference triplets, we pad the similarity matrix to size $k \times k$, where \( k = \max(m,n).\)
We then solve a bipartite matching problem over the binary assignment matrix
$
X \in \{0,1\}^{k \times k},
$
where $X_{ij}=1$ indicates that predicted triplet $\hat{\tau}_i$ is matched to reference triplet $\tau_j^*$. 
The optimal assignment is obtained by solving:
\begin{equation}
X^* =
\arg\max_{X}
\sum_{i=1}^{k}\sum_{j=1}^{k} S_{ij}X_{ij},
\end{equation}
subject to one-to-one matching constraints
\(
\sum_{j=1}^{k} X_{ij} \leq 1,
\qquad
\sum_{i=1}^{k} X_{ij} \leq 1.
\)
After obtaining the optimal assignment $X^*$, we compute the total valid matching score:
\begin{equation} 
\mathcal{S}_{\mathrm{HS}} =
\frac{1}{\max(m,n)}
\sum_{(i,j)\in\mathcal{M}} S_{ij},
\end{equation}
where $\mathcal{M}$ denotes the set of matched prediction-reference pairs satisfying $X_{ij}^*=1$ with $i\le{m}$ and $j\le{n}$. This score rewards predicted reasoning steps that are semantically aligned with the reference decomposition and grounded in the correct visual evidence.

\textbf{Reward Design.}
The total reward combines three components: a final-answer reward, a format reward, and a Hungarian reasoning reward ($\mathcal{R}_{\mathrm{HS}}$). We define the overall reward as
\begin{equation}
\mathcal{R}
=
\alpha\,
\mathcal{R}_{\mathrm{format}}
+
\mathcal{R}_{\mathrm{answer}}
\mathcal{R}_{\mathrm{HS}},
\label{eq:overall_reward}
\end{equation}
where $\alpha$ controls the contribution of the format reward. The format reward encourages the model to follow the required output structure and is applied independently of answer correctness. In contrast, the Hungarian reasoning reward is controlled by the final-answer reward through the product $\mathcal{R}_{\mathrm{answer}}\mathcal{R}_{\mathrm{HS}}$. Thus, semantically aligned and visually grounded intermediate reasoning is rewarded only when it leads to a correct final answer.
The answer reward is also modulated by the Hungarian reasoning reward.
This design encourages the model to produce well-formed outputs while prioritizing reasoning trajectories that are both correct at the outcome level and well aligned with the reference grounded reasoning process.

\textbf{Format Reward}
The format reward ensures that the model follows the required structured reasoning protocol. 
We check whether the completion contains paired sub-questions and sub-answers, an explicit reasoning path, and a final answer.
Let $\mathrm{count}(\cdot)$ denote the number of occurrences of a required tag pattern. 
The format reward is defined as
$
\mathcal{R}_{\mathrm{format}} =
\frac{1}{3}
\left(
\mathbb{I}_{\mathrm{pair}}
+
\mathbb{I}_{\mathrm{reason}}
+
\mathbb{I}_{\mathrm{final}}
\right).
$
Here, $\mathbb{I}_{\mathrm{pair}}$ verifies that every sub-question is paired with an intermediate answer and there is at least one sub-question:
$
\mathbb{I}_{\mathrm{pair}} =
\mathbf{1}\left[
{\scriptstyle (\mathrm{count}(\mathcal{T}_q)
=
\mathrm{count}(\mathcal{T}_a))
\;\land\;
\mathrm{count}(\mathcal{T}_q) > 0}
\right]
$.
$\mathbb{I}_{\mathrm{reason}}$ verifies the presence of an explicit reasoning path:
$
\mathbb{I}_{\mathrm{reason}} =
\mathbf{1}\left[
\mathrm{count}(\mathcal{T}_r) > 0
\right]
$
.
$\mathbb{I}_{\mathrm{final}}$ verifies the presence of a final answer:
$
\mathbb{I}_{\mathrm{final}} =
\mathbf{1}\left[
\mathrm{count}(\mathcal{T}_f) > 0
\right].
$

\textbf{Final Answer Reward} evaluates the correctness of the final answer extracted from the structured output
\[
\mathcal{R}_{\mathrm{answer}} =
\begin{cases}
1, & \text{if } \hat{a}_{\mathrm{final}} = a^*_{\mathrm{final}}, \\
0, & \text{otherwise}.
\end{cases}
\]

\textbf{Hungarian Reasoning Reward.}

We convert the Hungarian matching score into a dense process-level reward:
\[
\mathcal{R}_\text{HS} = \max(0, \mathcal{S}_\text{HS} - \gamma),
\]
where $\gamma$ is a threshold controlling the minimum acceptable grounded reasoning quality. This thresholding mechanism suppresses weak or poorly grounded reasoning trajectories while preserving positive reward for trajectories whose intermediate reasoning steps are sufficiently aligned with the reference decomposition. Consequently, the grounding reward is activated only when the generated reasoning demonstrates adequate semantic and grounding consistency.

\textbf{H-GRPO Optimization.}
For each input $(I,Q)$, H-GRPO samples a group of $G$ trajectories from the current policy. 
Each trajectory is scored using the composite reward $\mathcal{R}$ (Eq. \ref{eq:overall_reward}).
Group-level rewards are then used to compute relative advantages, following the GRPO-style optimization principle. Overall, H-GRPO optimizes the model to satisfy three conditions simultaneously:  output must follow the required reasoning format, final answer must be correct, and intermediate reasoning must be visually grounded and semantically aligned with reference reasoning steps.
\section{Experiments and Results}
\textbf{Training data and protocol.}
We construct a synthetic grounded reasoning dataset, denoted as 
Grounded Visual Reasoning Steps (\database) dataset, to provide supervision for both supervised fine-tuning and reinforcement learning. The dataset construction details are given in Appendix~\ref{appendix.sec.training.data}.
Training consists of two stages. 
First, we perform supervised fine-tuning for one epoch on~\database. 
This stage teaches the model to follow the required output format and provides an initial behavioural prior for decomposed grounded reasoning. 
Second, we apply GRPO-based reinforcement learning with $G=8$ rollouts per sample. Both stages use the same optimization hyperparameters: learning rate $5 \times 10^{-6}$, warm-up ratio $0.03$, and weight decay $0.01$. We include more details in Appendix~\ref{appendix.sec.training.details}.

\textbf{Baselines and VLM backbone.}
We evaluate our approach using Qwen2.5-VL-3B~\cite{bai2025qwen25vltechnicalreport} (\textbf{Default}) and Gemma-4-E2B~\cite{team2026gemma} as VLM backbones. We compare against the following baselines and ablations to study the effects of task-specific fine-tuning, reinforcement learning, grounded reasoning, and permutation-invariant step matching:
\begin{itemize}[leftmargin=*, itemsep=0pt, topsep=0pt]

\item \textbf{Zero-shot}: The pretrained VLM is evaluated on the QA task without task-specific fine-tuning.

\item \textbf{SFT}: The VLM is supervised fine-tuned on \database~ and subsequently evaluated on the QA task.

\item \textbf{GRPO-QA}: Starting from the SFT model, the VLM is LoRA-tuned using standard GRPO with rewards on output format and final-task correctness, without explicitly optimizing intermediate grounded reasoning.

\item \textbf{H-GRPO}: Our proposed method, which LoRA-tunes the VLM to optimize grounded intermediate reasoning using permutation-invariant Hungarian matching.

\item \textbf{H-GRPO-F}: An ablation of H-GRPO that uses the same reasoning representation and reward formulation but replaces Hungarian matching with fixed (identity) step matching, i.e.,
$R_{\mathrm{fixed}} = \frac{1}{n}\sum_i S_{ii}$.
This variant therefore assumes an order-sensitive one-to-one correspondence between predicted and reference reasoning steps.

\item \textbf{H-GRPO-NG}: An ablation of H-GRPO that removes the IoU-based spatial grounding reward (\emph{No Grounding}), allowing us to isolate the contribution of explicit visual grounding supervision.

\item \textbf{H-GRPO+}: A two-stage variant in which the VLM is first optimized using H-GRPO-F and subsequently refined using H-GRPO. The first stage encourages learning of structured, order-sensitive reasoning trajectories, while the second introduces permutation-invariant matching to accommodate alternative valid reasoning-step orderings.

\end{itemize}

\textbf{Evaluation benchmarks.}
We evaluate performance of our method out-of-domain on unseen datasets such as RoboSpatial~\cite{song2025robospatial}, 
CV-Bench~\cite{tong2024cambrian}, and EmbSpatial-Bench~\cite{du2024embspatial}.
The three benchmarks share a common class of object-centric visual spatial reasoning tasks in which a model must identify relevant entities and infer qualitative geometric relationships between them from a static visual observation. This setting is well aligned with our central hypothesis that explicitly grounding intermediate reasoning steps can improve spatial reasoning. 
\begin{table*}[t]
\centering
\caption{\textbf{Performance on RoboSpatial.}
Answer accuracy evaluates spatial reasoning performance.
For grounded reasoning, IoU measures spatial grounding against the ground-truth
bounding boxes (reported separately for Compatibility and Configuration subsets). Mean Max. $\mathrm{sim}_q$ and Mean Max.
$\mathrm{sim}_a$ denote the average semantic similarity between each predicted sub-question/sub-answer and its most similar ground-truth counterpart, respectively.
Grounding metrics are not available for the zero-shot model.}
\label{tab:robospatial_results}
\scriptsize
\begin{tabular}{l c ccc ccc}
\toprule
&
\textbf{Answer}
&
\multicolumn{3}{c}{\textbf{Compatibility}}
&
\multicolumn{3}{c}{\textbf{Configuration}}
\\
\cmidrule(lr){3-5}
\cmidrule(lr){6-8}
\textbf{Method}
&
\textbf{Acc.}
&
\textbf{IoU}
&
\shortstack{\textbf{Mean Max.}\\$\mathrm{sim}_q$}
&
\shortstack{\textbf{Mean Max.}\\$\mathrm{sim}_a$}
&
\textbf{IoU}
&
\shortstack{\textbf{Mean Max.}\\$\mathrm{sim}_q$}
&
\shortstack{\textbf{Mean Max.}\\$\mathrm{sim}_a$}
\\
\midrule
Zero-shot
& 49.30
& -- & -- & --
& -- & -- & -- \\

\midrule
SFT
& 64.43
& 0.1427  &	0.6160 &	0.5746
& 0.2705  &	0.6996 & 0.6839 \\

\midrule
GRPO-QA
& 69.38
& 0.1566 & 0.6165 & 0.5838
& 0.2680 & 0.6968 & 0.6872 \\

H-GRPO-F
& 70.28
& 0.1535 & 0.6328 & 0.5983
& 0.2891 & 0.7260 & 0.6995 \\

H-GRPO-NG
& \textbf{72.23}
&0.1539	 & 0.6272	& 0.5942	
& 0.2754	& 0.7224	& 0.6947 \\

H-GRPO
& 70.81
& 0.1643 & 0.6492 & 0.6008
& 0.2792 & 0.7396 & 0.7040 \\

H-GRPO+
& 67.82
& \textbf{0.1682} & \textbf{0.6507} & \textbf{0.6074}
& \textbf{0.2894} & \textbf{0.7411} & \textbf{0.7115} \\
\bottomrule
\end{tabular}
\end{table*}

\begin{center}
\begin{minipage}[t]{0.49\columnwidth}
\refstepcounter{table}
\label{tab:cvbench_results}
\textbf{Table \thetable: Performance on CV-Bench}

\vspace{3pt}

\resizebox{\linewidth}{!}{
\begin{tabular}{lcccc}
\toprule
\textbf{Method}
& \textbf{Acc.}
& \textbf{IoU}
& \shortstack{\textbf{Mean Max.}\\$\mathrm{sim}_q$}
&\shortstack{\textbf{Mean Max.}\\$\mathrm{sim}_a$}
\\
\midrule
Zero-shot
& 65.39 & -- & -- & -- \\
\midrule
SFT
& 66.83
& 0.2574 &	0.6432 &	0.6646 \\
\midrule
GRPO-QA
& 68.95 & 0.2639 & 0.6447 & 0.6656 \\
H-GRPO-F
& 69.37 & 0.2739 & 0.6559 & 0.6744 \\
H-GRPO-NG
& 69.22	& 0.2648	& 0.6531	& 0.6731 \\
H-GRPO
& 69.33 & 0.2812 & 0.6598 & 0.6726 \\
H-GRPO+
& \textbf{69.79}
& \textbf{0.2900}
& \textbf{0.6654}
& \textbf{0.6794} \\
\bottomrule
\end{tabular}
}
\end{minipage}
\hfill
\begin{minipage}[t]{0.49\columnwidth}
\refstepcounter{table}
\label{tab:embspatial_results}
\textbf{Table \thetable: Performance on EmbSpatial-Bench}

 \vspace{2.5pt}

\resizebox{\linewidth}{!}{
\begin{tabular}{lcccc}
\toprule
\textbf{Method}
& \textbf{Acc.}
& \textbf{IoU}
& \shortstack{\textbf{Mean Max.}\\$\mathrm{sim}_q$}
&
\shortstack{\textbf{Mean Max.}\\$\mathrm{sim}_a$}
\\
\midrule
Zero-shot
& 46.54 & -- & -- & -- \\
\midrule
SFT
& 47.14
& 0.2616	& 0.6156	& 0.6386 \\
\midrule
GRPO-QA
& 49.04 & 0.2752 & 0.6211 & 0.6437 \\
H-GRPO-F
& 50.85 & 0.3008 & 0.6403 & 0.6603 \\
H-GRPO-NG &
49.04	& 0.2782	& 0.6308 &	0.6506 \\
H-GRPO
& 51.51 & 0.3078 & 0.6424 & 0.6597 \\
H-GRPO+
& \textbf{52.34}
& \textbf{0.3177}
& \textbf{0.6527}
& \textbf{0.6704} \\
\bottomrule
\end{tabular}
}
\end{minipage}

\end{center}

\textbf{Out-of-domain performance and grounded reasoning.}
Tables~\ref{tab:robospatial_results}, \ref{tab:cvbench_results}, and \ref{tab:embspatial_results} evaluate both final-answer accuracy and grounded intermediate reasoning on unseen spatial and geometric reasoning benchmarks. SFT consistently improves over zero-shot performance, with the largest gain on RoboSpatial (49.30$\rightarrow$64.43), while GRPO-QA further improves accuracy across all three datasets. Introducing explicit intermediate reasoning through H-GRPO-F generally provides additional gains over direct-answer GRPO-QA, increasing accuracy from 69.38 to 70.28 on RoboSpatial, 68.95 to 69.37 on CV-Bench, and 49.04 to 50.85 on EmbSpatial-Bench. H-GRPO further improves performance on RoboSpatial and EmbSpatial-Bench while remaining comparable on CV-Bench, indicating that permutation-invariant optimization of the reasoning process can provide additional benefits beyond enforcing a fixed-order structured decomposition. H-GRPO+ achieves the highest accuracy on CV-Bench (\textbf{69.79}) and EmbSpatial-Bench (\textbf{52.34}), together with the strongest intermediate grounding and decomposition metrics: IoU/${\mathrm{sim}_q}$/${\mathrm{sim}_a}$ reach \textbf{0.2900}/\textbf{0.6654}/\textbf{0.6794} on CV-Bench and \textbf{0.3177}/\textbf{0.6527}/\textbf{0.6704} on EmbSpatial-Bench. A similar improvement in process quality is observed on RoboSpatial, where H-GRPO+ obtains the highest IoU, ${\mathrm{sim}_q}$, and ${\mathrm{sim}_a}$ on both Compatibility (\textbf{0.1682}/\textbf{0.6507}/\textbf{0.6074}) and Configuration (\textbf{0.2894}/\textbf{0.7411}/\textbf{0.7115}). Interestingly, RoboSpatial exhibits a different relationship between process quality and final-answer accuracy: H-GRPO-NG achieves the highest answer accuracy (\textbf{72.23}) despite removing the grounding reward, whereas H-GRPO+ achieves substantially stronger grounding but a lower accuracy of 67.82. Consistent with this distinction, H-GRPO-NG produces weaker grounding than H-GRPO+ across all six RoboSpatial grounding metrics. These results show that strong answer accuracy can be achieved without explicitly rewarding spatial overlap, but this does not yield the strongest grounded reasoning quality. Conversely, stronger alignment of intermediate reasoning with reference evidence does not necessarily translate into proportional gains in final answer accuracy. This motivates evaluating outcome-level correctness and process-level grounding jointly when assessing grounded spatial reasoning.

\textbf{Relationship between grounding quality and answer correctness.}
The preceding results show that our process-oriented methods generally improve the quality of intermediate grounded reasoning, but stronger grounding at the dataset level does not always coincide with higher final-answer accuracy. We therefore examine this relationship more directly at the individual-example level. For each test sample, we compute the Pearson correlation between final-answer correctness and two measures of intermediate reasoning quality: the mean IoU of the matched grounding boxes and the mean sub-answer similarity (${\mathrm{sim}_a}$). Since final-answer correctness is binary, this is equivalent to the point-biserial correlation. We additionally report $R_{\mathrm{IoU}}$ and $R_{{\mathrm{sim}_a}}$, defined as the mean grounding quality among correctly answered examples divided by the corresponding overall mean. A ratio above $1$ therefore indicates that correctly answered examples exhibit stronger-than-average grounding or sub-answer quality. Further details are provided in Appendix~\ref{sec.app.quality.metrics}.

Table~\ref{tab:grounding_correctness_correlation} reveals a dataset-dependent but informative relationship between intermediate reasoning quality and answer correctness. On CV-Bench, IoU is positively correlated with correctness for all methods ($r=0.1592$--$0.1902$), with all $R_{\mathrm{IoU}}>1$, indicating that correctly answered samples tend to have better localized evidence. The same pattern is even more consistent on EmbSpatial-Bench, where both IoU ($r=0.1545$--$0.1814$) and ${\mathrm{sim}_a}$ ($r=0.1741$--$0.2075$) are positively associated with correctness, and all corresponding ratios exceed $1$.
RoboSpatial exhibits a different behavior: IoU is weakly negative for all methods, while ${\mathrm{sim}_a}$ correlations remain near zero, with the largest positive association observed for H-GRPO+ (\(r=0.0649\)).
Together with the benchmark-level results, these observations indicate that better grounded intermediate reasoning is associated with correct answers on CV-Bench and EmbSpatial-Bench, but that the relationship is not universal: spatial overlap alone is not necessarily predictive of answer correctness on RoboSpatial.

\begin{table}[t]
\centering
\caption{\textbf{Sample-level relationship between grounded reasoning quality and final-answer correctness.}
}
\label{tab:grounding_correctness_correlation}
\scriptsize
\begin{tabular}{llcccc}
\toprule
\textbf{Dataset}
& \textbf{Method}
& $\mathbf{r(\mathrm{IoU},\,Correct)}$
& $\mathbf{R_{\mathrm{IoU}}}$
& $\mathbf{r({\mathrm{sim}_a},\,Correct)}$
& $\mathbf{R_{{\mathrm{sim}_a}}}$ \\
\midrule

\multirow{4}{*}{RoboSpatial}
& GRPO-QA   & \textbf{-0.0164} & 0.9708          & -0.0274  & 0.9959  \\
& H-GRPO-F  & -0.1484          & 0.9307          & -0.0361  & 0.9953  \\
& H-GRPO-NG & -0.0228          & \textbf{0.9996} &	0.0323  & 1.0048  \\
& H-GRPO    & -0.0606          & 0.9787          &  0.0090  & 1.0010 \\
& H-GRPO+   & -0.0355          & 0.9773          & \textbf{0.0649} & \textbf{1.0074} \\
\midrule

\multirow{4}{*}{CV-Bench}
& GRPO-QA   & 0.1592          & 1.0754          & 0.1134          & 1.0098 \\
& H-GRPO-F  & 0.1802          & 1.0824          & 0.1094          & 1.0093 \\
& H-GRPO-NG & 0.1636          & 1.0760	        & 0.1017          & 1.0084       \\
& H-GRPO    & 0.1756          & 1.0749          & \textbf{0.1210} & \textbf{1.0109} \\
& H-GRPO+   & \textbf{0.1902} & \textbf{1.0947} & 0.0972          & 1.0087 \\
\midrule

\multirow{4}{*}{EmbSpatial-Bench}
& GRPO-QA   & 0.1545          & 1.1105          & 0.1895          & 1.0329 \\
& H-GRPO-F  & 0.1623          & 1.1061          & 0.1860          & 1.0317 \\
& H-GRPO-NG & \textbf{0.1814} & \textbf{1.1320}	&\textbf{0.2075}  & \textbf{1.0363} \\
& H-GRPO    & 0.1705          & 1.1079          & 0.1950          & 1.0350 \\
& H-GRPO+   & 0.1600          & 1.0975          & 0.1741          & 1.0276 \\

\bottomrule
\end{tabular}
\end{table}

\textbf{Dependence on predicted visual evidence.}
Correlation alone does not establish whether the model actually uses the localized evidence when producing its final answer. We therefore perform a complementary intervention-based evaluation on H-GRPO+. For each sample, we mask the image regions covered by the bounding boxes predicted during the reasoning process and re-run the model on the resulting image using the same inference procedure. If these regions contain visual information used by the model for its final decision, removing them should degrade answer accuracy. Masking the predicted evidence produces substantial performance drops on all three benchmarks: from 67.82 to 56.90 on RoboSpatial ($-10.92$ points), from 69.79 to 49.51 on CV-Bench ($-20.28$ points), and from 52.34 to 37.14 on EmbSpatial-Bench ($-15.20$ points). Thus, the predicted grounding regions are not merely spatial annotations accompanying the reasoning output; removing their visual content substantially changes the model's ability to answer correctly. Together with the sample-level analysis above, this provides complementary evidence that the learned intermediate grounding identifies visual information relevant to the model's predictions.
\paragraph{Overall findings and discussion.}
Taken together, the experiments reveal several consistent observations. First, moving from direct-answer supervision toward explicit process-oriented optimization generally improves both spatial reasoning performance and the quality of intermediate reasoning. H-GRPO-F and H-GRPO typically improve over GRPO-QA, while H-GRPO+ achieves the strongest grounding and decomposition quality across all three benchmarks and the highest answer accuracy on CV-Bench and EmbSpatial-Bench. Second, fixed and permutation-invariant optimization provide complementary benefits: fixed matching encourages a structured reasoning trajectory, whereas Hungarian matching accommodates equivalent reasoning steps without imposing a predetermined ordering.

More importantly, our results reinforce a growing observation in recent VLM research that \emph{identifying relevant visual evidence and producing a correct answer are related but distinct capabilities}. Recent work on ``Seeing but Not Believing'' shows that VLMs can attend to relevant visual evidence yet still produce incorrect answers, revealing a disconnect between visual localization and effective evidence utilization~\cite{liu2026seeing}.
Similarly, EVICT~\cite{kotte2026evict} uses counterfactual evidence dropout---masking cited visual evidence and measuring the resulting change in the answer---to test whether a prediction actually depends on its purported evidence. Our results expose a similar distinction from the perspective of explicit grounded reasoning. On RoboSpatial, for example, H-GRPO-NG achieves the highest answer accuracy (\textbf{72.23}) despite removing the grounding reward, whereas H-GRPO+ produces substantially stronger grounding and decomposition quality. The sample-level analysis further shows that grounding quality is positively associated with correctness on CV-Bench and EmbSpatial-Bench, but only weakly so on RoboSpatial. 
We provide further qualitative analysis in Appendix~\ref{appendix.sec.qualitatives}.

Stronger grounding metrics alone do not establish that localized evidence is causally used for prediction. Our evidence-masking experiment addresses this by showing substantial accuracy drops across all three benchmarks when H-GRPO+ predicted regions are removed, indicating that these regions materially influence the model's answers. Together, our results support evaluating grounded VLM reasoning beyond answer accuracy, considering reasoning quality, spatial grounding, and dependence on identified visual evidence. H-GRPO+ jointly targets these process-level properties rather than final-answer accuracy alone.

\paragraph{Comparisons with other VLMs and other similar methods.}
Reason-RFT~\cite{tan2025reasonrft} and R1-VL~\cite{zhang2025r1vl} provide relevant compact-model baselines for spatial reasoning. On RoboSpatial, Reason-RFT-2B achieves an accuracy of \textbf{57.7}, while R1-VL-2B reports \textbf{52.5}. These results indicate that reinforcement-based post-training can substantially improve spatial reasoning in small VLMs. However, we did not find directly comparable reported results for Reason-RFT or R1-VL on CV-Bench and EmbSpatial-Bench under the same evaluation protocol, so we avoid making cross-benchmark comparisons for those datasets.
ViGoRL~\cite{sarch2025vigorl}, which explicitly incorporates visual grounding into reinforcement learning, reports \textbf{67.1\%} accuracy on RoboSpatial with a 3B model and \textbf{76.4\%} with a 7B model. These models use either a comparable 3B setting or a larger 7B setting and therefore provide useful reference points for grounded reinforcement learning on spatial reasoning. ViGoRL does not report directly comparable results on CV-Bench or EmbSpatial-Bench, so we restrict the comparison to RoboSpatial.
Results using Gemma-4-E2B are also reported in Tables~\ref{tab:robospatial_results_gemma},\ref{tab:cvbench_results_gemma} and \ref{tab:embspatial_results_gemma} in Appendix.


\section{Conclusion}
We presented a framework for learning and evaluating \emph{grounded intermediate reasoning} in vision-language models, where complex spatial reasoning is decomposed into sub-questions, sub-answers, and their supporting visual evidence. To train such reasoning without imposing a single fixed reasoning trajectory, we introduced H-GRPO, which uses permutation-invariant Hungarian matching to align predicted and reference reasoning steps and jointly optimize their semantic and spatial grounding quality. We further developed a scalable data synthesis pipeline for constructing grounded reasoning supervision and extended existing spatial reasoning benchmarks with annotations and metrics for evaluating the intermediate reasoning process.

Our experiments show that explicitly optimizing grounded reasoning can improve both final-answer performance and the quality of intermediate reasoning, with H-GRPO+ achieving the strongest overall grounding and decomposition quality across the evaluated benchmarks. Importantly, our analyses reveal that final-answer correctness and grounded process quality are related but distinct: stronger grounding is associated with correctness on several benchmarks, yet high answer accuracy does not necessarily imply well-grounded reasoning. Furthermore, masking the visual regions identified during reasoning substantially reduces answer accuracy, providing evidence that the predicted grounding captures information relevant to the model's decisions. These findings highlight the importance of evaluating not only whether a VLM reaches the correct answer, but also whether its intermediate reasoning is semantically meaningful and grounded in relevant visual evidence. We hope this framework provides a step toward VLMs whose reasoning processes are more explicitly grounded, measurable, and verifiable.

\subsection*{AI use statement}

In this work, we used generative AI tools for paper editing and table arrangements, literature survey, generate cleaning and reformatting the synthetic data and implementing methods.
We have not used generative AI tools for
proposing or refining hypotheses, designing or providing feedback on research methodology or experiments, assisting with translation, supporting qualitative and thematic data analysis, interpreting results.
Developing theoretical models or conceptual frameworks, formulating mathematical claims, providing critical ingredients for proving mathematical claims, and assisting in the writing of proofs, do not apply to this work.

Additionally, we used generative AI tools to
create or modify scientific figures or images, create or edit software code, draft parts of a research paper, summarize or analyse existing literature, brain-storming, sourcing/searching for information, edit a research paper to improve readability, identify relevant literature. 
We have reviewed all AI-assisted work. 
We inspected any generated figures were used only for drafting and were manually edited, refined or recreated, we verified and tested LLM-generated code for correctness. 
We take responsibility for the final content of this work, including text, claims or artifacts produced with the aid of generative AI.





\subsection*{Reproducibility statement}

We have taken several steps to facilitate the reproducibility of our results. Detailed experimental settings, implementation details, hyperparameter configurations, and additional information required to reproduce the reported experiments are provided in Appendix~\ref{appendix.sec.training.details}. We also provide descriptions of the datasets and creation process in Appendix~\ref{appendix.sec.training.data}.
The datasets and source code used to conduct our experiments will be released publicly to support independent verification and future research.



\subsubsection*{Acknowledgments}
This research/project is supported by the National Research Foundation, Singapore, under its NRF Fellowship Award No. NRF-NRFF14-2022-0001. This research is also supported by funding allocated to B.F. by the Agency for Science, Technology and Research (A*STAR) under its SERC Central Research Fund (CRF), as well as its Centre for Frontier AI Research (CFAR).

\bibliographystyle{plain}
\bibliography{iclr2027_conference}

\appendix
\section{Appendix}
\section{Datasets and models}
Our experiments utilize publicly released datasets obtained from HuggingFace where available or the respective works repositories. 
All resources were used in accordance with their licensing agreements and intended usage conditions.

\subsection{Training data.}
\label{appendix.sec.training.data}
We construct a synthetic grounded reasoning dataset, denoted as 
Grounded Visual Reasoning Steps (\database) dataset, to provide supervision for both supervised fine-tuning and reinforcement learning. 
Each training sample consists of an image $I$, a question $Q$, a final answer $a^{final}$, and a structured reasoning trace composed of intermediate triplets: $(q_i, a_i, b_i)$.
The dataset is generated through a multi-stage pipeline. 
First, we manually build a high-quality reference set of 40 representative visual reasoning examples. 
Each example is initially drafted using GPT-4o~\cite{hurst2024gpt} and then refined through human-in-the-loop verification to ensure  consistency and answer correctness. 
To improve spatial precision, we further refine the grounding annotations using SAM3~\cite{carion2025sam}, where generated sub-answers are used as textual prompts for box refinement. 
This reference set serves as the gold-standard set for prompt selection and reward calibration.
Next, we construct a pool of 12 candidate system prompts. 
Three LLMs, GPT-4o~\cite{hurst2024gpt}, Gemini-3~\cite{team2023gemini}, and Qwen3.5-Omni-Plus~\cite{qwenteam2026qwen35omnitechnicalreport}, are each asked to generate four candidate prompts using the reference examples. 
Each prompt is evaluated by generating grounded reasoning traces for the reference set and scoring them with our composite reward function. 
The best-performing prompt is then used to generate 10,000 training samples.
To encourage broad spatial reasoning coverage, \database\ is constructed from four sources: Visual7W~\cite{zhu2016visual7w}, Visual-CoT~\cite{shao2024visual}, A-OKVQA~\cite{schwenk2022okvqa}, and ERQA~\cite{team2025geminirobotics}. 
We manually validate the quality of the generated traces using three expert annotators on a random subset of 100 examples. 
The annotators evaluate consistency of sub-questions, semantic correctness of sub-answers, and spatial accuracy of bounding boxes using a custom annotation interface, described in Appendix~\ref{appendix:evaluator_interface}.

\subsection{Training Dataset Distribution}
\begin{figure}[h!]
    \centering
    \includegraphics[width=0.3\linewidth]{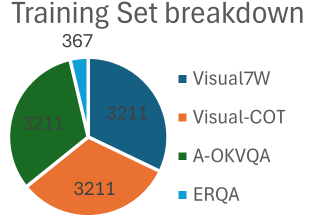}    
    \caption{Distribution of the composition of our 10,000-sample training set}    
    \label{fig:Training set distribution}
\end{figure}
For our training set, we show the distribution of samples from the original dataset in Figure ~\ref{fig:Training set distribution}. We randomly sampled instances from the training splits of the source datasets while maintaining approximately balanced contributions. However, owing to the limited size of ERQA, its representation is significantly smaller than that of the other datasets.
\begin{figure}[ht]
    \centering
    \includegraphics[width=\textwidth]{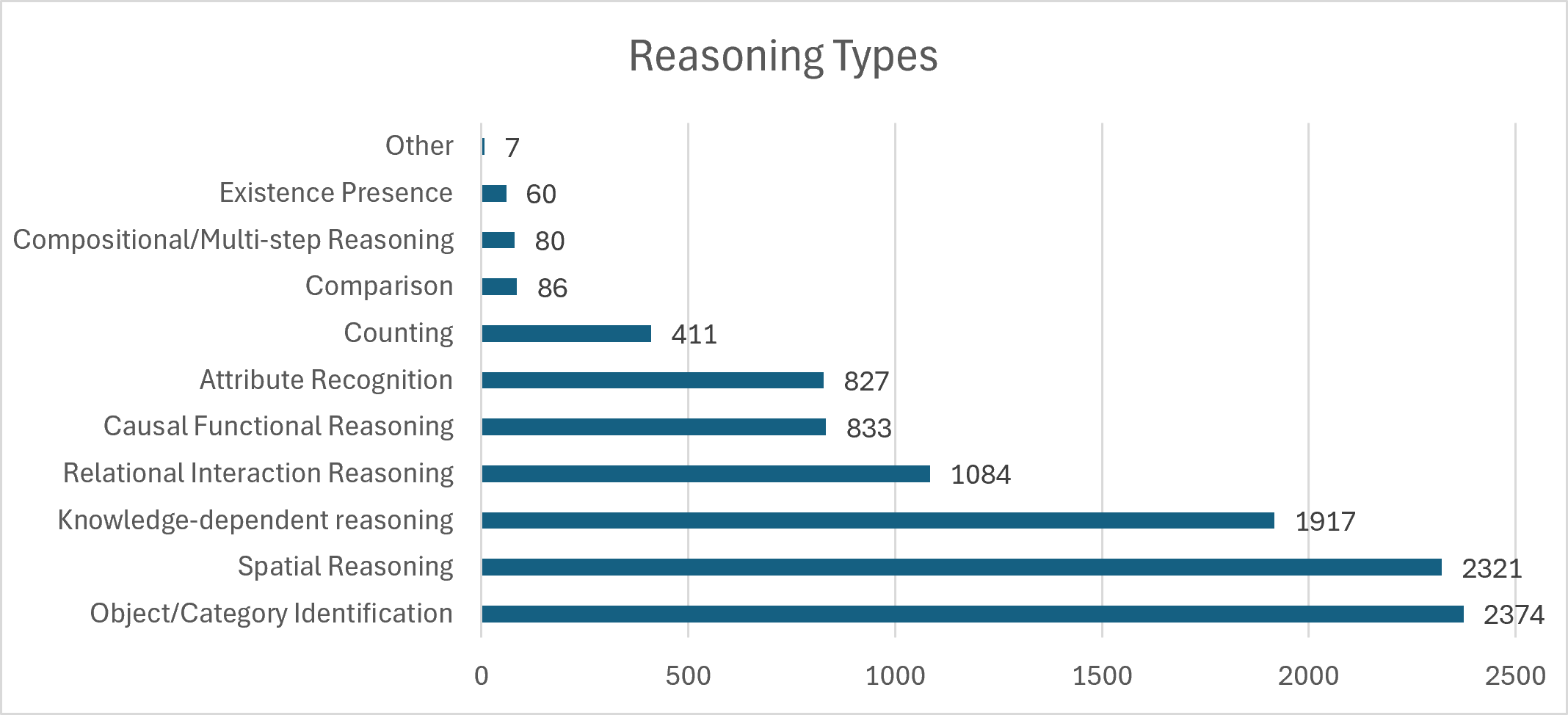}    
    \caption{Reason type composition in the dataset.}    
    \label{fig:Reasoning_Breakdown}
\end{figure}
The composition of the reasoning types in the dataset is shown in Figure~\ref{fig:Reasoning_Breakdown}.

\section{Training Details and Hyperparameters}
\label{appendix.sec.training.details}
We conduct our experiments with 4 × A40 48GB GPUs on a Linux distribution running
CUDA 12.8. We provide the versions of key packages and models used in the experiments in Table~\ref{appendix:training_packages}.
For inference, we utilize FlashAttention2~\cite{dao2022flashattention} and VLLM~\cite{kwon2023efficient} for acceleration.
We include training hyperparameters in Table~\ref{appendix:SFT_training_hyperparameters} and Table~\ref{appendix:GRPO_training_hyperparameters}.

\begin{center}

\begin{minipage}[t]{0.30\columnwidth}
\centering
\refstepcounter{table}
\label{appendix:SFT_training_hyperparameters}
\textbf{Table \thetable: SFT stage hyperparameters.}

\vspace{3pt}

\resizebox{\linewidth}{!}{
\begin{tabular}{lc}
\toprule
\textbf{Hyperparameter} & \textbf{Value} \\
\midrule
Seed & 42 \\
Epochs                  & 1 \\
Learning rate           & $5.00 \times 10^{-6}$ \\
Weight decay            & 0.01 \\
Warmup ratio            & 0.03 \\
Batch size              & 8 \\
Gradient accumulation   & 1 \\
Scheduler               & cosine \\
Precision               & bf16 \\
Flash attention         & FA2 \\
DeepSpeed               & ZeRO Stage 2 \\
\bottomrule
\end{tabular}
}
\end{minipage}
\hfill
%
\begin{minipage}[t]{0.30\columnwidth}
\centering
\refstepcounter{table}
\label{appendix:GRPO_training_hyperparameters}
\textbf{Table \thetable: GRPO stage hyperparameters.}

\vspace{3pt}

\resizebox{\linewidth}{!}{
\begin{tabular}{lc}
\toprule
\textbf{Hyperparameter} & \textbf{Value} \\
\midrule
Seed &42 \\
Steps                   & 2500 \\
Learning rate           & $5.00 \times 10^{-6}$ \\
Weight decay            & 0.01 \\
Warmup ratio            & 0.03 \\
Batch size              & 4 \\
Gradient accumulation   & 2 \\
Scheduler               & cosine \\
Precision               & bf16 \\
Flash attention         & FA2 \\
DeepSpeed               & ZeRO Stage 2 \\
Num. generations        & 8 \\
KL coefficient          & 0.04 \\
Reward $\alpha$           & 0.1 \\
Reward $\gamma$           & 0.5 \\
\bottomrule
\end{tabular}
}
\end{minipage}
\hfill
%
\begin{minipage}[t]{0.30\columnwidth}
\centering
\refstepcounter{table}
\label{appendix:training_packages}
\textbf{Table \thetable: Package versions used.}

\vspace{3pt}

\resizebox{\linewidth}{!}{
\begin{tabular}{lc}
\toprule
\textbf{Package} & \textbf{Version} \\
\midrule
Python               & 3.10.0 \\
Transformers         & 4.57.3 \\
TRL                  & 0.24.0 \\
PyTorch              & 2.8.0 \\
DeepSpeed            & 0.18.4 \\
Sentence-Transformers & 5.2.0 \\
FlashAttention       & 2.8.3 \\
vLLM                 & 0.10.2 \\
SentenceBert model  & all-MiniLM-L6-v2 \\
\bottomrule
\end{tabular}
}
\end{minipage}

\end{center}

\section{Human Evaluation Details}
\label{appendix:evaluator_interface}
\begin{figure*}[!htbp]
    \centering
    \includegraphics[width=\textwidth]{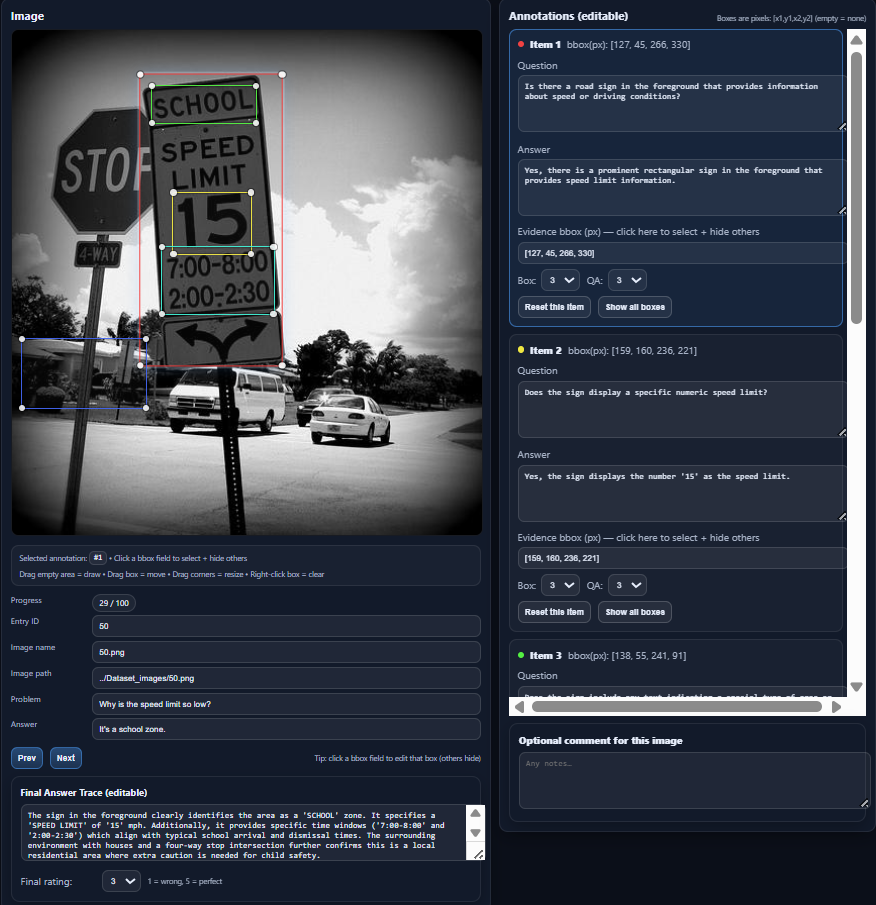}    
    \caption{Screenshot of the interface used in human evaluation of data quality. On the left side, from top to bottom: the image with bounding boxes overlaid, metadata, question and answer for the image, and the Final Answer Trace or Reasoning Path with a dropdown list for annotators to score the Final Answer Trace.\\
    On the right, is the sequential sub-question, sub-answers, and bounding boxes with a dropdown list for each triplet to score the quality of the QA pairs and bounding boxes.}
    \label{fig:Human_evaluation_interface}
\end{figure*}

\begin{tcolorbox}[         
    colback=white,      
    colframe=black,     
    boxrule=0.5pt,      
    sharp corners,      
    colupper=black,     
    boxsep=0.2pt,          
    fontupper=\small, 
    fontlower=\small
]
\textbf{Instructions to annotators} ``For each image you will be provided with the Main Question and Answer. The main question is broken down into a series of sub-questions and sub-answers on the right, with a bounding box where applicable to show the evidence used in answering the associated sub question.

Rank each sub question and answer pair on a scale of 1-5 based on relevance and correctness in helping to answer the main question.
Rank the bounding boxes on a scale of 1-5 based on how accurate and relevant it is to the sub-answers.

At the bottom left of the page, there is a Final Answer Trace. 
Rank the quality of the reasoning path on a scale of 1-5 based on the correctness and relevance in using the sub-questions and sub-answers to answer the main question.''
\label{Intructions_annotators}
\end{tcolorbox}

Data verification was performed by three expert annotators from the research team. No monetary compensation was provided. We include a brief summary of the instructions.

We include a screenshot in Figure~\ref{fig:Human_evaluation_interface} of the evaluation tool used by experts to verify the consistency of the generated data. Evaluators can utilise the interface to grade the QA pairs, bounding box accuracy and Final Answer Trace on a scale of 1-5.

\section{Results using Gemma-4-E2B}

\begin{table*}[t]
\centering
\caption{\textbf{Performance on RoboSpatial using Gemma-4-E2B.}
}
\label{tab:robospatial_results_gemma}
\resizebox{\textwidth}{!}{
\begin{tabular}{l c ccc ccc}
\toprule
&
\textbf{Answer}
&
\multicolumn{3}{c}{\textbf{Compatibility}}
&
\multicolumn{3}{c}{\textbf{Configuration}}
\\
\cmidrule(lr){3-5}
\cmidrule(lr){6-8}
\textbf{Method}
&
\textbf{Acc.}
&
\textbf{IoU}
&
\shortstack{\textbf{Mean Max.}\\$\mathrm{sim}_q$}
&
\shortstack{\textbf{Mean Max.}\\$\mathrm{sim}_a$}
&
\textbf{IoU}
&
\shortstack{\textbf{Mean Max.}\\$\mathrm{sim}_q$}
&
\shortstack{\textbf{Mean Max.}\\$\mathrm{sim}_a$}
\\
\midrule
Zero-shot
& 42.60
& -- & -- & --
& -- & -- & -- \\

H-GRPO-F
& \textbf{69.21}
& \textbf{0.2475}
& 0.6421
& 0.6098
& 0.4003
& 0.7229
& 0.6982 \\

H-GRPO+
& 68.35
& 0.2431
& \textbf{0.6591}
& \textbf{0.6188}
& \textbf{0.4088}
& \textbf{0.7274}
& \textbf{0.7025} \\
\bottomrule
\end{tabular}
}
\end{table*}
\begin{center}
\begin{minipage}[t]{0.49\columnwidth}
\centering
\refstepcounter{table}
\label{tab:cvbench_results_gemma}
\textbf{Table \thetable: Performance on CV-Bench using Gemma-4-E2B.}
\vspace{3pt}

\resizebox{\linewidth}{!}{
\begin{tabular}{lcccc}
\toprule
\textbf{Method}
& \textbf{Acc.}
& \textbf{IoU}
&
\shortstack{\textbf{Mean Max.}\\$\mathrm{sim}_q$}
&
\shortstack{\textbf{Mean Max.}\\$\mathrm{sim}_a$} \\

\midrule
Zero-shot
& 71.27 & -- & -- & -- \\

\midrule
H-GRPO-F
& 71.04 
& 0.4041
& 0.6706
& 0.6874 \\

H-GRPO+
& \textbf{71.49}
& \textbf{0.4200}
& \textbf{0.6729}
& \textbf{0.6890} \\
\bottomrule
\end{tabular}
}
\end{minipage}
\hfill
\begin{minipage}[t]{0.49\columnwidth}
\centering
\refstepcounter{table}
\label{tab:embspatial_results_gemma}
\textbf{Table \thetable: Performance on EmbSpatial-Bench using Gemma-4-E2B.}
\vspace{3pt}

\resizebox{\linewidth}{!}{
\begin{tabular}{lcccc}
\toprule
\textbf{Method}
& \textbf{Acc.}
& \textbf{IoU}
&
\shortstack{\textbf{Mean Max.}\\$\mathrm{sim}_q$}
&
\shortstack{\textbf{Mean Max.}\\$\mathrm{sim}_a$}\\
\midrule
Zero-shot
& 43.27 & -- & -- & -- \\

\midrule
H-GRPO-F
& 52.55
& 0.3813
& 0.6432
& 0.6612 \\

H-GRPO+
& \textbf{53.38}
& \textbf{0.3967}
& \textbf{0.6454}
& \textbf{0.6645} \\
\bottomrule
\end{tabular}
}
\end{minipage}
\end{center}

\section{Relationship between grounding quality and answer correctness.}
\label{sec.app.quality.metrics}
We report the ratio between the mean grounding quality of correctly answered examples and the corresponding overall mean grounding quality, denoted by $R_{\mathrm{IoU}}$ and $R_{\mathrm{sim}_a}$. Specifically,

\[
R_{\mathrm{IoU}} =
\frac{\mathbb{E}[\mathrm{IoU}\mid\mathrm{Correct}]}
{\mathbb{E}[\mathrm{IoU}]},
\qquad
R_{{\mathrm{sim}_a}} =
\frac{\mathbb{E}[\mathrm{sim}_a\mid\mathrm{Correct}]}
{\mathbb{E}[\mathrm{sim}_a]}.
\]
A ratio greater than $1$ indicates that correctly answered examples exhibit stronger-than-average visual grounding or sub-answer semantic alignment. Together, the correlation and normalized ratios provide complementary measures of the relationship between intermediate grounded reasoning quality and final-answer correctness.

There is a direct relation.
Let
\[
X=\mathrm{IoU},\qquad C\in\{0,1\},
\]
and define
\[
\mu_1=E[X\mid C=1],\quad
\mu_0=E[X\mid C=0],\quad
p=P(C=1),
\]
with overall mean
\[
\mu=E[X]=p\mu_1+(1-p)\mu_0.
\]
Our IoU ratio is
\[
R_{\mathrm{IoU}}=\frac{\mu_1}{\mu}.
\]
For binary \(C\), the Pearson correlation (point-biserial correlation) is
\[
r
=
\frac{(\mu_1-\mu_0)\sqrt{p(1-p)}}{\sigma_X}.
\]
Because
\[
\mu_1-\mu
=(1-p)(\mu_1-\mu_0),
\]
we obtain
\[
\mu_1-\mu
=
r\,\sigma_X\sqrt{\frac{1-p}{p}}.
\]
Therefore,
\[
\boxed{
R_{\mathrm{IoU}}
=
1+
\frac{r\,\sigma_{\mathrm{IoU}}}{\mu_{\mathrm{IoU}}}
\sqrt{\frac{1-p}{p}}
}
\]
where \(p\) is answer accuracy.
So there is indeed a direct relation:
\[
\boxed{
R_{\mathrm{IoU}}-1
\propto r
}
\]
for fixed accuracy \(p\), IoU mean \(\mu\), and IoU standard deviation \(\sigma\).

\section{Interpretability Evaluation}
Because existing methods do not emit the same structured triplet format, they cannot be evaluated directly using our grounded reasoning protocol, we evaluate the interpretability of generated reasoning traces using Gemini 3 Flash~\cite{googledeepmind2025gemini3flash} as an LLM-as-a-Judge. 
Each model output is scored on five dimensions: relevance, coherence, consistency, clarity, and overall interpretability, using a $1$--$5$ scale. 
The evaluation prompt is provided in Appendix~\ref{appendix:interpretability}.
As shown in Table~\ref{Intepretability}, H-GRPO achieves the highest score across all dimensions. 
It obtains an overall interpretability score of $4.73$, compared with $4.25$ for R1-VL, $3.81$ for ViGoRL, and $3.02$ for Reason-RFT. 
It also achieves the highest clarity score of $4.92$ and relevance score of $4.94$.
These results show that H-GRPO produces more interpretable reasoning traces than the compared methods.
The improvement is particularly clear in coherence and consistency, where H-GRPO scores $4.77$ and $4.74$, respectively. 
This suggests that grounding each intermediate answer in a localized region helps the model produce reasoning chains that are easier to inspect and verify. 
Instead of generating an unstructured textual chain-of-thought or a single final prediction, H-GRPO produces a sequence of visually anchored reasoning steps.

\begin{table}[h]
\small
  \centering
  \caption{LLM-as-a-Judge evaluation of reasoning interpretability. 
Interp.: overall interpretability, Rel.: relevance, Coh.: coherence, Cons.: consistency, and Clr.: clarity. Scores are measured on a $1$--$5$ scale, where higher is better.}
  \label{Intepretability}
    \begin{tabular}{lccccc}    
    \toprule
          \textbf{Method} & \textbf{Interp.} & \textbf{Rel.} & \textbf{Coh.} & \textbf{Cons.} & \textbf{Clr.} \\
            \midrule
            Reason-RFT   & 3.02 & 3.80 & 3.12 & 3.31 & 3.97 \\
            ViGoRL       & 3.81 & 4.60 & 3.85 & 4.04 & 4.25 \\
            R1-VL        & 4.25 & 4.84 & 4.33 & 4.34 & 4.79 \\
            H-GRPO       & \textbf{4.73} & \textbf{4.94} & \textbf{4.77} & \textbf{4.74} & \textbf{4.92} \\
    \bottomrule
    \end{tabular}     
\end{table}

\section{Qualitative Analysis}
\label{appendix.sec.qualitatives}

\begin{figure*}[t]
    \centering
    \includegraphics[width=\linewidth]{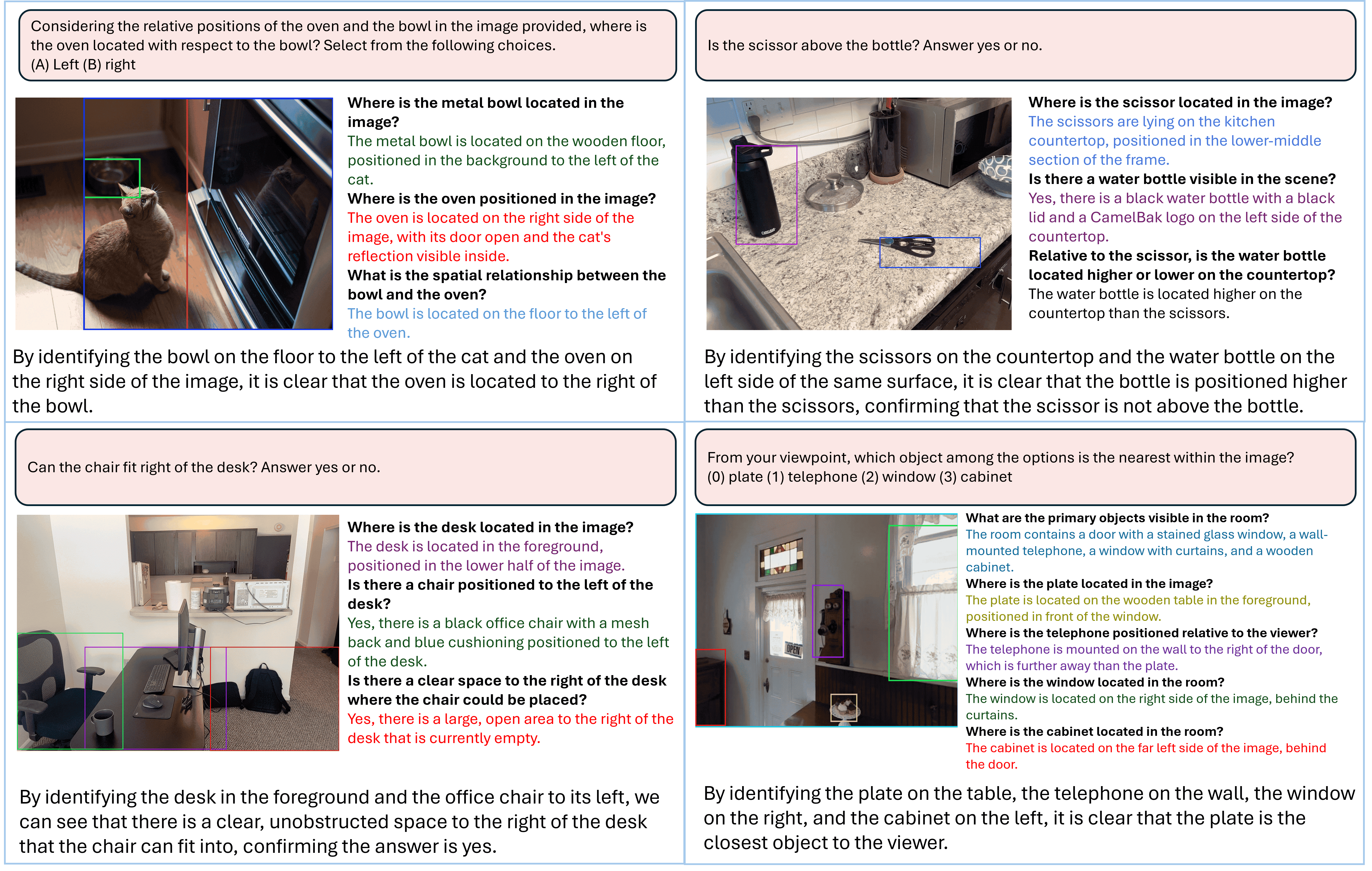}
    \caption{\textbf{Representative qualitative examples of spatially grounded reasoning on CV-Bench, RoboSpatial, and EmbSpatial.}
For each image-question pair, H-GRPO decomposes the prediction into intermediate steps that identify and localize relevant entities before reasoning about their spatial relationships.
The examples cover relative position and viewpoint-dependent relations such as left/right, above/below, and nearest-object reasoning.
By explicitly grounding both the relevant objects and supporting visual evidence in the image, H-GRPO produces an interpretable reasoning path from visual evidence to the final answer.}
    \label{fig:Qualitative}
\end{figure*}

\subsection{Representative Qualitative analysis}
Figure~\ref{fig:Qualitative} illustrates the spatially grounded reasoning behavior of H-GRPO across examples from CV-Bench, RoboSpatial, and EmbSpatial.
The examples demonstrate several forms of spatial reasoning.
In relative-position questions, the model separately localizes the objects before determining relations such as left/right or above/below.
For instance, when reasoning about the position of an oven relative to a bowl in the example on the top left panel of Figure~\ref{fig:Qualitative}, the model first identifies the bowl and the oven and then compares their image locations to infer that the oven lies to the right of the bowl.
Similarly, for the scissors-and-bottle example on the top right, it grounds both objects on the countertop before determining their vertical ordering.
The chair example on the bottom left further shows how the model combines object localization with free-space reasoning to determine whether a chair can fit within a specified region.
The qualitative results also highlight more challenging viewpoint-dependent and reference-based reasoning.
In the room scene, H-GRPO identifies multiple candidate objects and their locations before selecting the object that is spatially nearest from the specified viewpoint.
Such examples require the model to account for the image coordinate frame, object-relative relations, and the observer's perspective rather than relying only on object recognition.

The intermediate decomposition exposes which entities are considered, where they are localized, and how their spatial relationships contribute to the final answer.
This is particularly useful for spatial reasoning benchmarks, where a correct final prediction can otherwise obscure errors in object correspondence, reference-frame interpretation, or relational reasoning.
By exposing and grounding these intermediate decisions, H-GRPO provides a more inspectable account of how spatial conclusions could be derived from visual evidence.

\begin{figure*}[t]
    \centering
    \includegraphics[width=\linewidth]{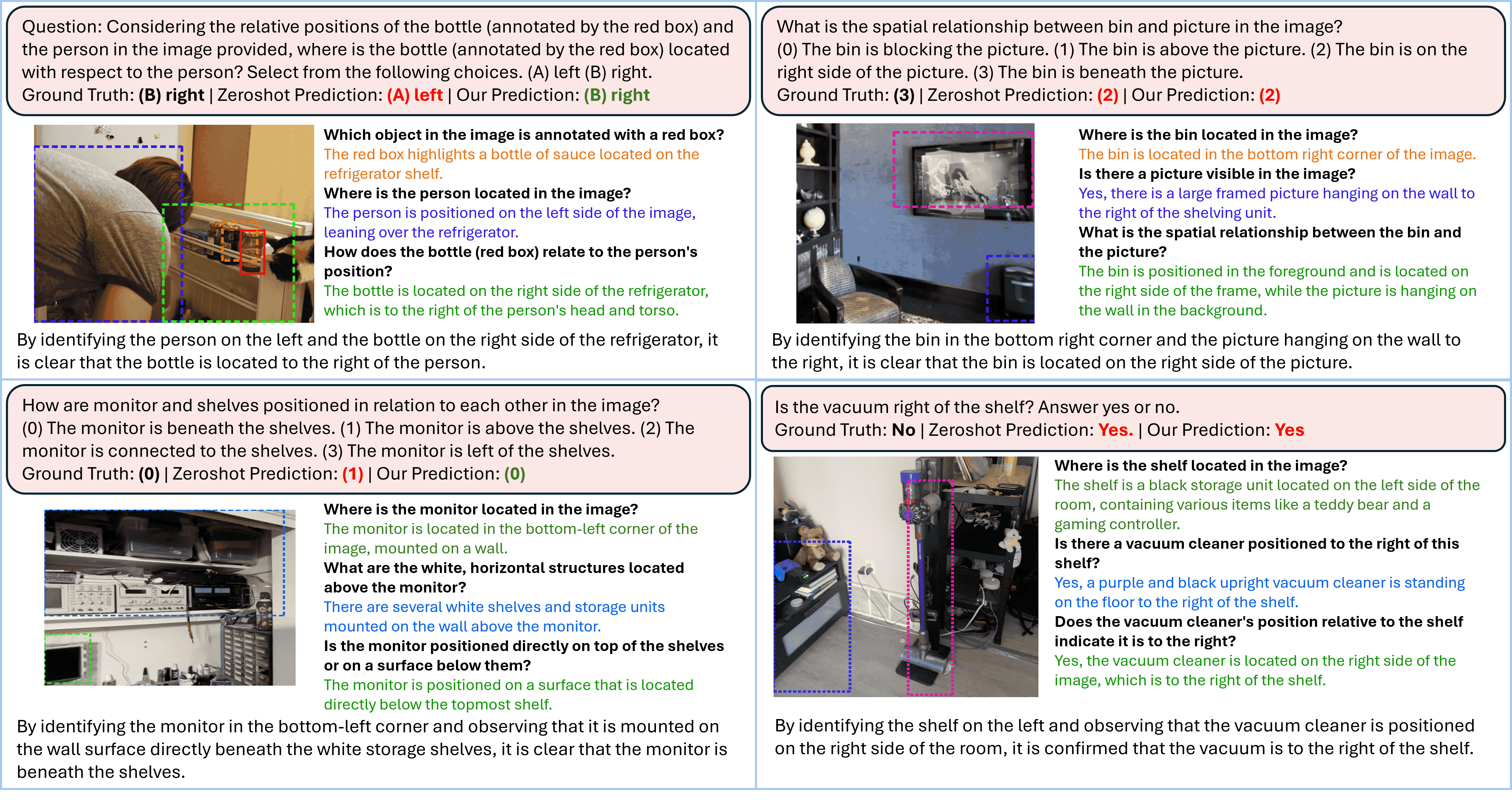}
    \caption{\textbf{Qualitative comparison between zero-shot and H-GRPO predictions.} Each example shows the full input question, ground-truth answer, zero-shot prediction, and H-GRPO prediction, together with H-GRPO’s grounded intermediate reasoning steps. The examples include cases where grounded decomposition corrects a zero-shot spatial error as well as cases where both models remain incorrect, illustrating both the benefits and limitations of the proposed approach.}
    \label{fig:Qualitative_zeroshot_comparison}
\end{figure*}

\subsection{Comparison with zero-shot predictions.}
Figure~\ref{fig:Qualitative_zeroshot_comparison} provides a targeted comparison between the pretrained zero-shot model and H-GRPO.
For each example, we report the ground-truth answer, the zero-shot prediction, and the H-GRPO prediction, together with the intermediate grounded reasoning generated by H-GRPO.
This comparison highlights how explicit decomposition and visual grounding can change the model's spatial reasoning behavior relative to the pretrained model.

In the top left panel Figure~\ref{fig:Qualitative_zeroshot_comparison} , the task asks for the position of the bottle relative to the person. The zero-shot model predicts the incorrect left-right relation, whereas H-GRPO first localizes the relevant entities and then correctly infers that the bottle lies to the right of the person. A similar pattern is observed in the  monitor-and-shelves (bottom left) example. The zero-shot model incorrectly predicts that the monitor is above the shelves, while H-GRPO decomposes the question, grounds both entities, and correctly identifies the monitor as being below the shelves.

The comparison also exposes cases in which explicit grounding does not lead to a correct final answer. In the bin-and-picture (top right) and vacuum-and-shelf example (bottom right), H-GRPO produces the same incorrect final prediction as the zero-shot model, but its intermediate decomposition makes the source of the error easier to inspect. In the vacuum-and-shelf example, the reasoning trace shows that the model grounds the TV console on the left as the relevant shelf, rather than the shelf located to the right of the vacuum. In the bin-and-picture example, the intermediate trace reveals the spatial interpretation that leads to the incorrect choice, showing that the final prediction is consistent with the model's interpretation even though that interpretation does not match the ground truth.

These examples reinforce the distinction between process-level grounding quality and final-answer correctness. Grounded intermediate reasoning can make the basis of a prediction more transparent and help localize the source of an error, but it does not guarantee that the selected visual referents or the resulting spatial inference are correct.

\subsection{Failure case analysis}
\begin{figure*}[!htbp]
    \centering
    \includegraphics[width=.95\textwidth]{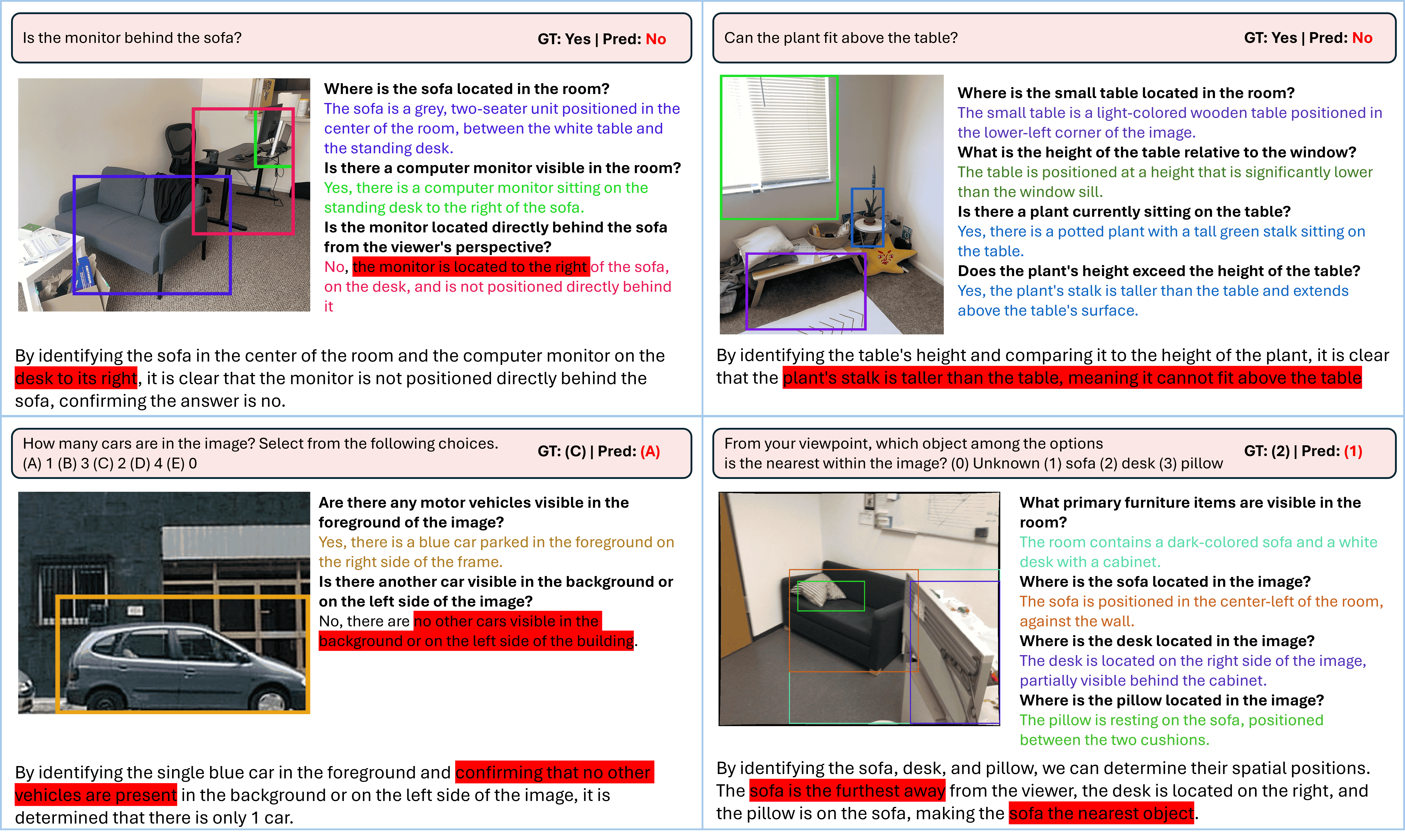}    
    \caption{Representative failure cases in spatially grounded reasoning, with erroneous reasoning steps highlighted in red. 
    \textbf{Top left:} incorrect interpretation of the spatial relation.
    \textbf{Top right:} failure in geometric/free-space reasoning. 
    \textbf{Bottom left:} counting failure due to missed partial object grounding. 
    \textbf{Bottom right:} self-contradictory spatial reasoning.}
    \label{fig:failure_cases}
\end{figure*}
Figure~\ref{fig:failure_cases} highlights representative failure modes of H-GRPO on spatial reasoning tasks. These examples show that errors can occur even when the relevant entities are correctly localized, indicating that spatial grounding alone does not guarantee correct relational reasoning.

In the top-left example, the model identifies both the sofa and the monitor but misinterprets the relation behind, instead focusing on the monitor being to the right of the sofa. The top-right example similarly shows correct object grounding but incorrect geometric reasoning, although the plant and table are localized correctly, the model incorrectly concludes that the plant cannot fit above the table.

The bottom-left example illustrates a grounding related counting failure, where the model detects the prominent car but misses a partially visible second vehicle at the image boundary. In the bottom-right example, the model correctly identifies and localizes the relevant objects, but its reasoning becomes internally inconsistent. It states that the sofa is the furthest object from the viewer, yet subsequently concludes that the sofa is the nearest object. This self-contradictory reasoning constitutes an internal inconsistency in the intermediate reasoning trace, where the final conclusion is not supported by the model's own preceding spatial assessment.

Overall, these cases suggest that H-GRPO's errors could arise from both incomplete visual grounding and incorrect reasoning over otherwise correctly grounded entities. The decomposition makes these failure modes explicit, allowing errors in perception, spatial relation interpretation, and final inference to be distinguished in the final prediction.

\clearpage

\section{Generation prompt}
\label{appendix:Generation_prompt}
\begin{tcolorbox}[
    breakable,
    colback=white,      
    colframe=black,     
    boxrule=0.5pt,      
    sharp corners,      
    colupper=black,     
    boxsep=.2pt,          
    fontupper=\small, 
    fontlower=\small
]
\textbf{Prompt for Generation}``Objective \\
You are an expert Visual Question Answering (VQA) and grounding model. Your task is to validate and break down a given Question and its corresponding Answer about an Image. You must decompose the main question into a series of logical, sequential, and increasingly specific Sub-Questions that, when answered, collectively and progressively lead to the final provided Answer. \\
Instructions for Output Generation \\
Analyze the Input: Carefully consider the provided Image, the main Question, and the final Answer. \\
Decomposition Strategy: \\
Break the main Question down into a minimum of three and a maximum of six distinct Sub-Questions. \\
The sequence of Sub-Questions must follow a logical path: \\
Start General: Begin with general, high-level observations about the image relevant to the main Question's topic (e.g., "What objects are present?"). \\
Progress to Specific: Gradually narrow the focus to specific details, relationships, attributes, and actions required to confirm or refute the final Answer (e.g., "What is the color of the object?", "How is object A positioned relative to object B?"). \\
Avoid Cheating: Sub-Questions must not pre-emptively reveal the final Answer or state information that is clearly only part of the final conclusion. \\
Answering and Grounding: \\
Provide a concise Answer for each Sub-Question. \\
Generate a single evidence\_bbox (bounding box) for each sub-question's Answer. \\
The bounding box must ground the specific visual evidence that supports the sub-question's answer. \\
Bounding box coordinates must be normalized from 0 to 1000 (e.g., [x\_min, y\_min, x\_max, y\_max]). \\
If the answer requires a global observation or the absence of an element, set evidence\_bbox to null. \\
Final Path: \\
Create a final\_answer\_path which is a single sentence or brief paragraph that logically synthesizes the observations from all the Sub-Questions to explain how they collectively prove the final Answer. \\
Example Input and Required JSON Output \\
Input Provided (Imagine these are the variables you receive): \\
Image: [The provided image] \\
Question: "How is the traffic pattern controlled?" \\
Answer: "with solar power" \\
REQUIRED JSON OUTPUT FORMAT: \\
JSON \\
\{ \\
  "sub\_questions": [ \\
    \{ \\
      "question": "Is there a visible device that appears to power or control traffic?", \\
      "answer": "Yes, there is a solar panel mounted on a pole.", \\
      "evidence\_bbox": [40, 40, 140, 180]  \\
    \}, \\
    \{ \\
      "question": "Is there any indication of electrical wiring or external power sources?", \\
      "answer": "No, no wired power infrastructure is visible.", \\
      "evidence\_bbox": null \\
    \}, \\
    \{ \\
      "question": "Is the solar panel positioned in a way that suggests it powers signage or signals?", \\
      "answer": "Yes, the solar panel is attached above the stop sign, indicating it powers the control system.", \\
      "evidence\_bbox": [30, 110, 160, 260] \\
    \}, \\
    \{ \\
      "question": "Is there any other traffic-control equipment present?", \\
      "answer": "A stop sign is present, but the only power-related element is the solar panel.", \\
      "evidence\_bbox": [60, 150, 120, 240] \\
    \} \\
  ], \\
  "final\_answer\_path": "These observations show that the only visible power/control mechanism is a solar panel, which indicates that the traffic pattern is controlled with solar power." \\
\} \\

Your Input and Task \\
Image:\{image\_here\} \\
Question: \{question\} \\
Answer: \{answer\} \\
Generate the JSON object based on the Image, Question, and Answer following all the above instructions.''
\end{tcolorbox}

\section{Interpretability prompt}
\label{appendix:interpretability}
\begin{tcolorbox}[
    breakable,          
    colback=white,      
    colframe=black,     
    boxrule=0.5pt,      
    sharp corners,      
    colupper=black,     
    boxsep=0.5pt,          
    fontupper=\small, 
    fontlower=\small
]
\textbf{Prompt for interpretability} ``You are evaluating the interpretability of the following reasoning chain produced by an AI model for a visual reasoning question.

**Question:** \{question\} \\
**Ground-truth answer:** \{gt\} \\
**Model's reasoning chain:** \\

\{formatted\_prediction\} \\

---
Rate the interpretability of the reasoning chain above on a scale of 1 to 5:\\

1 - Completely uninterpretable: reasoning is incoherent, self-contradictory, or irrelevant. \\
2 - Mostly uninterpretable: some steps make sense but the overall chain is confusing or misleading. \\
3 - Partially interpretable: the logic is partially clear but has noticeable gaps or inconsistencies. \\
4 - Mostly interpretable: the reasoning is clear and mostly coherent with only minor issues. \\
5 - Fully interpretable: each step is clear, logically connected, and directly supports the final answer. \\

Consider the following dimensions: \\
- **Relevance**: Are the sub-questions/steps directly relevant to answering the main question? \\
- **Coherence**: Do the sub-answers/steps logically follow from the sub-questions/steps and build toward the conclusion? \\
- **Consistency**: Is the reasoning path consistent with the sub-answers/steps and does it support the final answer? \\
- **Clarity**: Is the language clear and unambiguous? \\

Respond ONLY with valid JSON in this exact format: \\
\begin{verbatim}
{
  "score": <integer 1-5>,
  "relevance": <integer 1-5>,
  "coherence": <integer 1-5>,
  "consistency": <integer 1-5>,
  "clarity": <integer 1-5>,
  "justification": "<one or two sentences explaining the score>"
}
\end{verbatim}
\end{tcolorbox}

\end{document}